\pdfoutput=1

\documentclass[11pt]{article}

\usepackage[]{acl}

\usepackage{times}
\usepackage{latexsym}
\usepackage{comment}
\usepackage[T1]{fontenc}

\usepackage[utf8]{inputenc}

\usepackage{microtype}

\usepackage{graphicx}
\usepackage[export]{adjustbox}
\usepackage{mdframed}

\title{XDBERT: Distilling Visual Information to BERT from Cross-Modal Systems to Improve Language Understanding
}

\author{Chan-Jan Hsu$^{1,2}$, Hung-yi Lee$^{1}$, Yu Tsao$^{2}$ \\
  $^{1}$National Taiwan University, Taiwan \\
  $^{2}$Academia Sinica, Taiwan\\
  \texttt{ \{r09946011, hungyilee\}@ntu.edu.tw,} \\ \texttt{yu.tsao@citi.sinica.edu.tw} 
}

\begin{document}
\maketitle
\begin{abstract}
Transformer-based models are widely used in natural language understanding (NLU) tasks, and multimodal transformers have been effective in visual-language tasks. This study explores distilling visual information from pretrained multimodal transformers to pretrained language encoders. Our framework is inspired by cross-modal encoders' success in visual-language tasks while we alter the learning objective to cater to the language-heavy characteristics of NLU. After training with a small number of extra adapting steps and finetuned, the proposed XDBERT (cross-modal distilled BERT) outperforms pretrained-BERT in general language understanding evaluation (GLUE), situations with adversarial generations (SWAG) benchmarks, and readability benchmarks. We analyze the performance of XDBERT on GLUE to show that the improvement is likely visually grounded.
\end{abstract}

\section{Introduction}
Transformer-based models are extensively used in natural language understanding (NLU) tasks, and some prominent pretraining strategies include BERT \citep{devlin-etal-2019-bert}, RoBERTa \citep{liu2019roberta}, ALBERT \citep{DBLP:conf/iclr/LanCGGSS20}, and ELECTRA \citep{DBLP:conf/iclr/ClarkLLM20}. Despite their differences in curating the learning objectives, they all utilize text-based datasets only. 
In the real world, however, humans can benefit from the visual modality when acquiring knowledge from language; an obvious example is learning visually grounded words, such as colors and shapes. 

Some studies have succeeded with visually grounded information used in NLU. ViCo~\citep{DBLP:conf/iccv/GuptaSH19} learned visual co-occurrences in text and reported superior performance to GloVe in word analogy problems. \citet{DBLP:conf/iclr/0001C0USLZ20} and \citet{huang-etal-2020-unsupervised-multimodal} used images to boost translation performance in supervised and unsupervised settings.  \citet{tan-bansal-2020-vokenization} reported improvements over BERT on NLU by proposing the concept of vokenization. 

\begin{figure}
    \centering
    \includegraphics{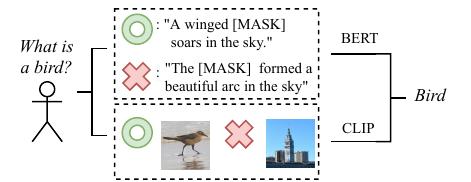}
    \caption{Humans can answer cloze questions and match a word with an image, and the multi-views of a word could be simulated by neural networks. While BERT excels in masked word reconstruction, CLIP (Section 3) specializes at image-text matching.  The two modalities have different collocations of concepts, which incentivize joint learning from the two systems. }
    \label{fig:multiview}
\end{figure}

Another branch of research focuses on solving multimodal downstream tasks such as visual question answering and image retrieval. \citet{DBLP:journals/corr/abs-1908-03557,  DBLP:conf/nips/LuBPL19, DBLP:conf/iclr/SuZCLLWD20, DBLP:conf/eccv/Li0LZHZWH0WCG20} trained visual-text transformers, while LXMERT \citep{tan-bansal-2019-lxmert} used different encoders for text and image and a cross-modal encoder. \citet{tan-bansal-2020-vokenization} tested these models with general language understanding evaluation (GLUE
 \citet{wang2018glue}) and found that the performance does not exceed using BERT (Appendix~\ref{sec:A}), drawing the conclusion that vision-and-language pretraining on visually-grounded language dataset
failed to distill useful information for general NLU. CLIP \citep{DBLP:journals/corr/abs-2103-00020} utilizes contrastive loss to reach SOTA on zero-shot image classification in a retrieval fashion. 

In this work, we establish the link between pretrained multimodal transformers and visually-grounded language learning. We devise a way to distill visual information from components of a pretrained multimodal transformer (CLIP text-transfomer, abbreviated as CLIP-T) to pretrained language transformers (BERT/ELECTRA), to incorporate versatile perception of words into the model (Figure ~\ref{fig:multiview}). 
The usage of a visually grounded text-transformer as a teacher allows us to implement straightforward and non-fuzzy adapting tasks for distillation. We show that it is mathematically logical that the CLIP-T output approximates visual features (Sec. ~\ref{sec:Pretraining}), and also the linguistic competence of CLIP-T is low (Sec. ~\ref{sec:Experimental Results}), to prove that the distilled information is predominantly visual and thus non-trivial to the pretrained-language transformer despite having textual inputs.

Methodologically, we use the cross-modal encoder structure inspired by \citet{tan-bansal-2019-lxmert}, to concatenate the two models and further adapt the ensemble for some extra steps (a lot fewer than the original pretraining steps). While adapting pretrained-BERT, we favor a document-level corpus (wiki103) over a vision-language corpus (MSCOCO) due to claims from \citet{devlin-etal-2019-bert}\footnote{"It is critical to use a document-level corpus rather than a shuffled sentence-level corpus such as the BillionWord Benchmark in order to extract long contiguous sequences"} and results from \citet{tan-bansal-2020-vokenization} (Appendix ~\ref{sec:A}). The adapting tasks are joint masked language modeling (MLM), same sentence prediction, and CLIP token classification tasks, which are resemblant of BERT pretraining tasks to cater to the language-heavy characteristics of NLU. We do ablation studies to show that each of the task provides improvement (Section ~\ref{sec:Ablation}).

During finetuning, we finetune XDBERT (cross-modal distilled BERT), which is the language encoder after adaptation. We evaluate the linguistic capabilities of the model by finetuning on GLUE, situations with adversarial generations (SWAG \citep{zellers2018swag})  benchmarks, and readability benchmarks\footnote{https://www.kaggle.com/c/commonlitreadabilityprize}. The resulting XDBERT outperforms pretrained BERT, proving that our adaptation strategy distills useful visual knowledge into BERT (right of Figure~\ref{fig:modelpipeline}). We provide analysis to show that the improvements are visually grounded. 

We summarize our contribution as follow:
\begin{itemize}
\item We explore distilling visual information from a pretrained multimodal transformer to a pretrained language transformer and improved NLU performance.
\item Our adapting method is efficient and extensible to different combinations of pretrained-language encoders (BERT/ELECTRA).
\end{itemize}

\section{Proposed Method}
\label{sec:Training}

\begin{figure*}
    \centering
    \includegraphics{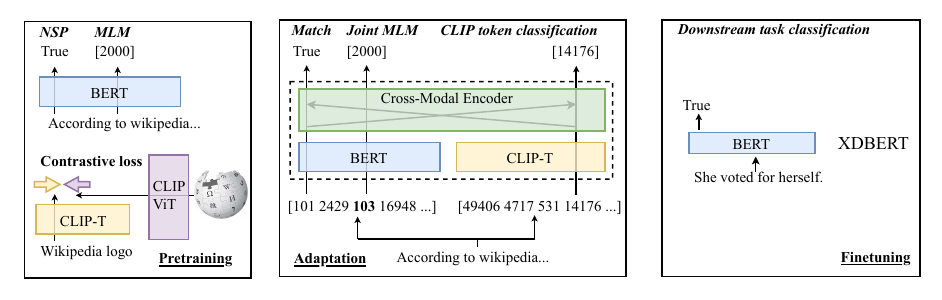}
    \caption{In our experimental setting, the transformers go through three phases of the training processes from left to right. The pretraining phase pretrains BERT and CLIP-T, both of which are then used in the adaptation phase and concatenated with a cross-modal encoder. Finetuning is performed on the language encoder only (XDBERT); in this case, a positive CoLA example  is being processed to determine its linguistic acceptability.  ViT stands for Vision Transformer \citep{dosovitskiy2021an}, and the input id 103 is the [MASK] token in BERT.}
    \label{fig:modelpipeline}
\end{figure*}

The training process consists of three phases: pretraining, adaptation, and finetuning (Figure~\ref{fig:modelpipeline}). Our proposed method focuses on the adaptation phase with pretrained models, so pretraining is not a part of our experiment, but we explain all three phases for completeness. The adaptation phase incorporates the cross-modal transformer structure to jointly learn from CLIP-T and BERT outputs. 

\subsection{Model Architecture}
\label{sec:Architecture}
\begin{figure}
    \centering
    \includegraphics{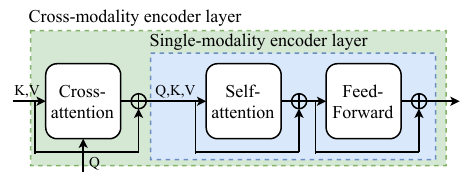}
    \caption{Single-modality encoder layer (blue) and cross-modal encoder layer (green)}
    \label{fig:cross-modal encoder}
\end{figure}

The cross-modal transformer (middle of Figure~\ref{fig:modelpipeline}) consists of a cross-modal encoder, CLIP-T and BERT. CLIP-T has the same module connections as BERT with only parameter differences (specifications in Appendix~\ref{sec:AppendixCLIPSequence}). The cross-modal encoder consists of repeating cross-modal encoder layers, which is an extension to single-modality encoder layers (layers of BERT/CLIP-T)  in Figure~\ref{fig:cross-modal encoder}. The added cross-attention module follows the attention formula \citep{DBLP:conf/nips/VaswaniSPUJGKP17}: 
\begin{equation} 
Attention\ output = softmax\left({\textbf{Q}*\textbf{K}^T}\slash{\sqrt{D}}\right)\textbf{V} 
\end{equation} 
for queries (\textbf{Q}), keys (\textbf{K}) and values (\textbf{V}) of dimension D, however, \textbf{Q} is generated from a modality other than \textbf{K} and \textbf{V}. We choose the number of cross-modal encoder layers to be 2.

\subsection{Pretraining}
\label{sec:Pretraining}
BERT is trained using the next sentence prediction and masked language modeling. CLIP is an image-text matching system with two components, a text encoder (CLIP-T), and an image encoder (CLIP-ViT), which learn to encode paired inputs to closer output embeddings via contrastive loss. The trained representation has the following properties:
\begin{equation} 
 cos(H_i, V_i) >> cos(H_i, V_j)  (i \neq j )
\end{equation}
\begin{equation} 
 cos(H_i, V_i) >> cos(H_j, V_i)  (i \neq j )
 \end{equation}

where $H_i$ is the CLIP text encoder output of $X_i$, and $V_i$ is the CLIP image encoder output of $Y_i$. The text-image input ($X_i$, $Y_i$) is paired, and every ($X_j$, $Y_k$) $(j \neq k )$ is a non-pair. Since $H_i$ and $V_i$ are normalized and have a length of 1, $H_i$ can be used to approximate $V_i$. The similarity of $H_i$ and $V_i$ is also shown in multi-modal arithmetic propreties discovered in \citet{DBLP:journals/corr/abs-2111-14447}
Therefore, we use the CLIP text encoder output to approximate CLIP image encoder output for a straightforward adaptation process. 
\subsection{Adaptation}
We define three adapting tasks that can be learned in a self-supervised manner, which is visualized in Figure~\ref{fig:modelpipeline}. In these tasks, BERT and CLIP-T takes sentences A and B respectively as input, and losses are calculated from both BERT output and CLIP-T output. Our adapting tasks closely follow BERT text pretraining strategies to retain linguistic competence. Unlike pretraining, the adaptation is computationally inexpensive, as we found that training 1 epoch on wiki103 was already effective. Further training details can be found in Appendix~\ref{sec:AppendixTrainingDetails}. 

\subsubsection{Joint Masked Language Modeling (MLM)}
The MLM objective teaches the model to reconstruct masked tokens. The masked ratio and masked token replacement probabilities follow \citet{devlin-etal-2019-bert}. 
 Since there is no equivalent of a [MASK] token in CLIP, we leave the sentence as is. 
\subsubsection{Same sentence prediction (MATCH)}
 The Image-Text Matching (ITM) objective is widely used in multimodal learning \citep{tan-bansal-2019-lxmert}. We modify this objective to same sentence prediction as both streams of our model takes text as input. When choosing the input sentences for BERT and CLIP-T, we make the inputs nonidentical 50\% of the time. A binary classifier over [CLS] differentiates between the two cases. This motivates the [CLS] output to encode sentence related information, and trains the cross-attention weights. 
\subsubsection{CLIP Token Classification} 
This is the MLM objective done on the CLIP-T side of the full model, omitting the masking part because CLIP has no mask token. Same as MLM, 15\% of the tokens are randomly selected for reconstruction. We address concerns on trivial solutions learned by the model in Section ~\ref{sec:Ablation} and ~\ref{tab:AttentionMap} in the appendix. 

\subsection{Finetuning} 
Finetuning follows the methods described in \citet{devlin-etal-2019-bert}, and is applied to the language encoder only (XDBERT), therefore the number of parameters are kept equal to pretrained-BERT. 
\begin{table*}
\centering
\small
\begin{tabular}{ccccccccccc}
\hline
\textbf{} &\textbf{RTE} & \textbf{MPRC} & \textbf{STSB} & \textbf{CoLA}  & \textbf{SST2} & \textbf{QNLI} & \textbf{QQP} & \textbf{MNLI} &   \textbf{SWAG} & \textbf{READ↓}  \\
\hline
CLIP-T & 51.62  & 76.20  & 22.07 & 25.41 &  -- &   -- &  -- &  -- & -- & --\\
\hline
BERT-b & 66.43  & 87.38  & 88.64 & 56.52 &  92.46 &  90.92 & 89.51 &  84.35 & 81.0 & --\\
XDBERT-b & \textbf{69.31}  & \textbf{88.02} & \textbf{89.32} & \textbf{57.55} & \textbf{92.78} & \textbf{91.52} &  \textbf{89.57} & \textbf{84.75} & \textbf{81.35}& --\\
\hline
ELECTRA-b & 78.70 & 89.49 & 90.77 & 66.09 & 94.5 & 92.69 & 90.29 & 88.23 &88.60& --\\
XDELECTRA-b & \textbf{80.51} & \textbf{90.55 }& \textbf{91.04} & \textbf{66.76} & \textbf{95.20} & \textbf{93.03} & \textbf{90.4} & \textbf{88.75} & \textbf{88.73} & -- \\
\hline
ELECTRA-l & 86.64 & 91.53 & 91.88 & 69.27 & 96.90 & 94.78 & \textbf{91.34} & 90.99 & 92.46 & 0.685\\
XDELECTRA-l & \textbf{87.73} & \textbf{92.12} & \textbf{91.97} & \textbf{70.98} & \textbf{97.36} & \textbf{94.93} & 91.29 & \textbf{91.02} & \textbf{92.59} & \textbf{0.635} \\
\hline
\end{tabular}
\caption{
NLU task results on the test set (READ) and the dev set (GLUE,SWAG). The results are the median value of 5 runs using different random seeds (9 runs on RTE). BERT-b is the BERT-base-uncased model from \citet{devlin-etal-2019-bert}, while XDBERT-b is the proposed models shown in the right part of Figure~\ref{fig:modelpipeline}. ELECTRA-b and ELECTRA-l refer to the ELECTRA-base model and the ELECTRA-large model from \citet{DBLP:conf/iclr/ClarkLLM20} respectively. READ (readability benchmark) uses RMSE loss as the evaluation metric.
}
\label{Results}
\end{table*}

\section{Experimental Results}
\label{sec:Experimental Results}
We evaluated our model on three NLU benchmarks, namely GLUE, SWAG and READ. We tested our adaptation strategy on three different language encoders coupled with CLIP-T, including BERT-base, ELECTRA-base, and ELECTRA-large. We fix the finetuning parameters between models where comparison is intended, and select the median result of multiple runs. Details of finetuning are provided in Appendix~\ref{sec:AppendixTrainingDetails}.

Table~\ref{Results} shows experimental results. Each of our XD-model constantly outperforms the original encoder (For fair comparison, we train the original encoder with one more epoch of wiki103). We found that performance gains are more significant on smaller datasets (RTE, MRPC, STSB, CoLA), indicating that visual features help increase generalization when the amount of training data is limited. The gains are also significant on the readability benchmark (READ).

We show that the results of finetuning CLIP-T alone on GLUE does not perform well. Since the language capability of the CLIP-T model is weak, the distilled information obtained by XDBERT/XDELECTRA is predominantly visual. 

It is also possible to finetune the entire cross-modal transformer after adaptation. The performance further increases but the model has more parameters. The results are in  Appendix~\ref{sec:AppendixXFinetune}.

\section{Analysis}
 To justify the use of a cross-modal encoder, we first conducted a pairwise projection weighted canonical correlation analysis (PWCCA) on word embeddings. The PWCCA is a good measure to determine how close the distributions of two vector groups are to each other. The PWCCA results in Table~\ref{tab:PWCCA} show low scores on both BERT/CLIP and ELECTRA/CLIP before co-training, so the cross-modal encoder is useful in learning from both distributions. 

\begin{table}[h]
\centering
\begin{tabular}{ccc}
    \textbf{Systems} &\textbf{PWCCA} \\
    \hline
   BERT/ELECTRA & 0.5498 \\
    BERT/CLIP & 0.4980 \\
   ELECTRA/CLIP  & 0.4645 \\
   BERT/RANDOM & 0.3569 \\
\end{tabular}
\caption{
PWCCA results for different combinations of systems. RANDOM denotes embeddings generated from a uniform distribution.}
\label{tab:PWCCA}
\end{table}
\begin{figure}[h]

    \includegraphics[left]{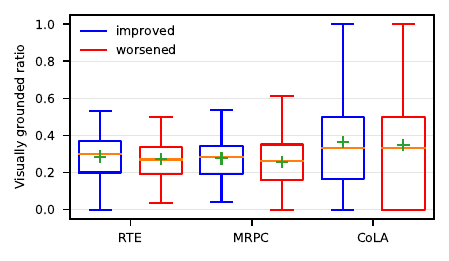}
    \caption{Characteristic analysis of RTE, MRPC, and CoLA entries categorized by performance difference between XDBERT-b and BERT-b. The Green plus symbol denotes the mean value. The visually grounded ratio estimation follows \citet{tan-bansal-2020-vokenization}.}
    \label{fig:analysis}
\end{figure}
We inspect RTE, MRPC, and CoLA results of 5 runs in detail to show that the improvements are likely from visual information of CLIP-T. Over the 5 runs, XDBERT-b has accumulated +38 more correct classifications than BERT-b, or +2.74\%(38/5/277) gain in performance. MPRC and CoLA show +0.3\% and +0.9\% gains in accuracy respectively, and translates to a larger gain in performance with their original metric (MRPC F1: +0.83\%, CoLA Corr: +2.2\%).
\begin{table*}
\centering

\begin{tabular}{ccccc}
\hline
\textbf{} &\textbf{RTE} & \textbf{MPRC} & \textbf{STSB} & \textbf{CoLA} \\
\hline
MLM+MATCH+CLIPTC(proposed) & 69.31  & 88.02 & 89.32 &  56.27 \\
MLM+MATCH & 70.04  & 86.93 & 88.8 &  54.62 \\
MLM & 68.23 & 87.25 & 89.29 & 54.78 \\
1 cross attention layer & 66.79  & 87.66 & 89.32 & 53.62 \\
\hline
2 Epochs (2x) & 69.31  & 88.04 & 89.31 & 55.91\\
20 Epochs (20x) & 57.4  & 87.74 & - & - \\
wiki(14G), same steps as above& 65.3  & 87.78 & 89.1 & -
\\
\hline
\end{tabular}

\caption{Ablation study results. The results are the median value of 5 runs using a learning rate of 1e-4 on XDBERT-b. The CoLA learning rate differs from that in the main paper.}
\label{tab:ablation}
\end{table*}
We then separate each of the glue datasets entries into two categories: entries that XDBERT-b improves classification over BERT-b, and entries of the opposite. Entries where both models obtain the same performance are set aside. Analyzing the separated entries as a whole, we discovered that the better-performing entries have a larger visually grounded ratio (Figure~\ref{fig:analysis}), as the quartile, median and mean values are generally higher for improved samples. The enhancement of visually grounded token representations is a rough indicator that XDBERT has obtained distilled visual information from CLIP-T. We show examples of each category in Appendix~\ref{sec:AppendixRTEExamples}.

\section{Ablation study}
\label{sec:Ablation}
We tried various combinations of adaptation tasks and found out that using all three yielded the best results. We also tried to reduce the number of cross-modal encoder layers to one; however, no further improvements were made upon the visually grounded language encoder. Other experiments include changing the number of layers in the cross-modal encoder, training for longer, and swapping to a much larger wiki (14G). Swapping to wiki reduces potential overfitting from the 20 Epochs setting trained on wiki103, as training for the same amount of steps on wiki is less than 1 epoch. We tested these changes on RTE, MPRC, STSB, and CoLA on 5 random seeds, and the results are shown in Table~\ref{tab:ablation}, where MLM refers to the joint MLM objective, MATCH refers to the cross-modal matching objective, and CLIPTC refers to the CLIP token classification objective. 

Besides experimental evidence, we also justify the CLIPTC loss via further analysis, as the CLIPTC objective can theoretically be trivially solved by identity mapping. Despite this possibility, we find that the loss is crucial to cross attention learning. Since we do not impose negative hard samples from sampled sentences, the MATCH objective can be solved sufficiently simply by guiding the cross attention to focus on common trivial words. With the CLIPTC objective, the diversity of the input embeddings corresponding to different tokens must be retained in the cross-modal encoder, leading to more robust cross-modal attention. We show comparisons of the attention maps generated from the cross-modal encoders with a random sequence from RTE in Table ~\ref{tab:AttentionMap} in the Appendix to verify this claim.
\section{Conclusion}
In this study, we explored using cross-modal encoders to distill visual information to BERT. We adapted the model with multiple objectives, and we were able to achieve improved performance on NLU tasks. Our adaptation techniques are computationally inexpensive and straightforward. Furthermore, our method is language encoder agnostic, as we show similar performance gains on XDELECTRA.

\section*{Acknowledgements}
This work was supported in part by Ministry of Science and Technology (MOST), Taiwan, under Contract 110-2223-E-002 -007-MY3. We thank the National Center for High-performance Computing (NCHC) of National Applied Research Laboratories (NARLabs) in Taiwan for providing computational and storage resources.

\bibliography{anthology,custom}
\bibliographystyle{acl_natbib}

\appendix

\section{Visual-Text Transformers Results on NLU}

\label{sec:A}
We show the result of Visual-Text Transformers on GLUE, reported by \citet{tan-bansal-2020-vokenization} in Table~\ref{tab:VisualModelsOnGlue}. All of the listed methods (except LXMERT) have their text-transformers initialized from BERT. The results show that multi-modal training for solving vision-language tasks does not improve the performance of the models on natural language understanding tasks.

\section{Modeling sequences on CLIP}
\label{sec:AppendixCLIPSequence}
While BERT and CLIP have similar forwarding mechanisms, the specifications of the transformer architecture are different, resulting in challenges to jointly model both models (Table~\ref{tab:specs}).
\begin{table}
\centering
\begin{tabular}{cccc}
\hline
\textbf{} & \textbf{BERT-b} & \textbf{BERT-l} & \textbf{CLIP}\\
\hline
dim & 768 & 1024 & 512 \\
max\_len & 512 & 512 & 77 \\
\#layers & 12 & 24 & 12 \\
\hline
\end{tabular}
\caption{BERT and CLIP configurations. ELECTRA has a structure identical to that of BERT. The tokenizers of BERT and CLIP are also different.}
\label{tab:specs}
\end{table}

Mismatching dimensions pose a problem in cross-attention. We use a linear transformation to generate \textbf{Q}, \textbf{K}, and \textbf{V} of matching dimensions, but clarify that this linear transformation layer exists in the original LXMERT setting where hidden representations have unified dimensions.

We modify the input to address the mismatched max\_len of the two systems. In the joint MLM, we used a fixed sequence length of 512 for the BERT. However, the same cannot be done for CLIP as the maxmum model sequence length is 77 for CLIP. We found that most BERT sequences (>99\%) of length 512 encode into CLIP sequences of length less than 693, so we pad the CLIP sequence to length 693, and then split the CLIP sequence into 9 sub-sequences of length 77. Therefore, a batch of inputs will contain BERT inputs of size (batch\_size, 512) and CLIP inputs of size (batch\_size, 9, 77). The output was resized to (batch\_size, 693) in the cross-modal encoder. The issue is also present in the finetuning phase, and the maximum sequence length of GLUE and SWAG is 128; therefore we used 2 blocks of CLIP sub-sequences to model it. For bi-sequence classification tasks such as RTE and MRPC, we ensure that separate sentences do not use the same block in the CLIP encoder. Therefore, uni-sequence classification tasks will have a CLIP input size of (batch\_size, 2, 77) and the bi-sequence classification task will have a CLIP input size of (batch\_size, 4, 77).

\section{Further Training Details}
\label{sec:AppendixTrainingDetails}
\subsection{Adaptation}
\label{sec:AppendixAdaptaion}
 We use publicly available wiki103  and preprocessing methods similar to \citet{tan-bansal-2020-vokenization} \footnote{\url{https://github.com/airsplay/vokenization}}. Wiki103 (500MB) is a subset of the Wikipedia corpus consisting of only good and featured articles. The adaptation of 1 epoch on wiki103 finished in 35 minutes on 8 V100s (BERT-base). We trained for at most 20 epochs(~16k steps) and found that further adaptation steps did not increase scores in early epochs, and significantly decreased performance in late epochs. We used the following parameters for adaptation : learning rate = 1e-4,  max\_epoch = 40 (although we stopped early due to plummeting performance), warmup ratio = 0.05

 \subsection{Finetuning}
 The learning rates are listed in Table~\ref{tab:finetuneconfig}. 
 \begin{table}[h]
\centering
\begin{tabular}{ccc}
\hline
\textbf{} & \textbf{base-sized} & \textbf{large-sized} \\
\hline
RTE,MRPC,STSB & 1e-4 & 5e-5  \\
others & 2e-5 & 1e-5  \\
\hline
\end{tabular}
\caption{Finetuning configurations for NLU tasks. The full model uses the same learning rate as its language encoder}
\label{tab:finetuneconfig}
\end{table}

We used a warmup ratio of 0.1, with a learning rate decay of 0.9, and trained the model for 3 epochs. We report the median results of 5 runs on different random seeds, except for RTE, which is unstable; therefore, we report the median results of 9 runs instead. The reproduce results of ELECTRA on RTE and STSB are lower than values reported by \citet{DBLP:conf/iclr/ClarkLLM20} because we did not start from an MNLI checkpoint.

 \subsection{Finetuning with Full Model}
 \label{sec:AppendixXFinetune}
 Since our cross-modal transformer itself is can also be viewed as a language encoder, finetuning can be done on the full model. This approach, however, adds extra parameters to pretrained-BERT, so comparison with pretrained-BERT is not intended, instead, we focus on showing the feasibility of this approach. The number of additional parameters is only a function of the hidden size in BERT/ELECTRA, so when the language encoder is large, the ratio of additional parameters is much more insignificant. To simplify notations, we use X-(language encoder) to represent the full model. The number of parameters of the full model is shown in Table~\ref{numofparameters} and the results on NLU tasks are shown in Table~\ref{tab:XBERT}.

\begin{table}[h]
\centering
\begin{tabular}{cc}
\hline
\textbf{model} & \textbf{parameters}\\
\hline
BERT-b / ELECTRA-b & 109482240 \\
XBERT-b / XELECTRA-b & 202059009 \\
ELECTRA-l & 334092288 \\
XELECTRA-l & 442671617 \\
\hline
\end{tabular}
\caption{Number of paramters for each model.}
\label{numofparameters}
\end{table}

\section{RTE Examples}

\label{sec:AppendixRTEExamples}
We provide three RTE example of each type in Figure\ref{fig:analysis}, and we choose extreme examples where performance difference is huge over 5 runs for both "Improved" and "Worsened" categories. We follow \citet{tan-bansal-2020-vokenization} to classify tokens as visually-grounded if it is not a stopword and has more than 100 occurrences in MSCOCO. In the following examples, \textbf{Bold} words are visually-grounded, while normal words are non-visually-grounded. Words in brackets are stopwords and does not count towards either category.
\subsection{Improved : XDBERT outperforms BERT}
Example1 : \\
Visually-grounded ratio : 11/(11+16) = 0.4074\\
BERT answered correctly : 0/5\\
XDBERT answered correctly : 5/5 

\begin{mdframed}
 \textbf{hands} \textbf{across} (the) divide (was) formed (in) march 2001 (,) (and) \textbf{one} (of) (its) immediate aims (was) (to) press (for) (more) freedom (of) contact (and) communication \textbf{right} \textbf{away} (between) (the) \textbf{two} \textbf{parts} (of) cyprus (,) (and) (for) early \textbf{progress} \textbf{towards} (a) solution (to) (') (the) cyprus problem (') (.)
 
 cyprus (was) divided (into) \textbf{two} \textbf{parts} (in) march 2001 (.)
\end{mdframed}
Example2 :\\ 
Visually-grounded ratio : 4/(10+4) = 0.2857\\
BERT answered correctly : 0/5\\
XDBERT answered correctly : 5/5 

\begin{mdframed}
(it) (is) hoped (that) \textbf{women} (,) (who) constitute (more) (than) \textbf{half} (of) (the) population (,) (will) vote (for) (other) \textbf{women} (and) ensure (that) (their) issues (are) represented (in) parliament (.) 

\textbf{women} (are) poorly represented (in) parliament (.)
\end{mdframed}
Example3 : \\
Visually-grounded ratio : 13/(13+17) = 0.4333\\
BERT answered correctly : 0/5\\
XDBERT answered correctly : 5/5 

\begin{mdframed}

\textbf{ho} \#\#dler claimed (there) (were) \textbf{also} irregularities (in) (the) campaigns \textbf{organized} (by) atlanta (for) (the) 1996 summer \textbf{games} (,) sydney (for) (the) summer olympics (in) 2000 (and) salt \textbf{lake} \textbf{city} (for) (the) 2002 \textbf{winter} \textbf{games} (.) 

(before) salt \textbf{lake} \textbf{city} (,) \textbf{winter} olympic \textbf{games} took \textbf{place} (in) naga \#\#no (.)
\end{mdframed}
\subsection{On Par : XDBERT and BERT perform equally}

Example1 : \\
Visually-grounded ratio : 6/(6+32) = 0.1375\\
BERT answered correctly : 0/5\\
XDBERT answered correctly : 0/5 
\begin{mdframed}
(on) october 1 2001 (,) eu (and) (other) countries introduced (the) option (for) domestic \textbf{animal} owners (to) apply (for) \textbf{pet} passports (under) (the) pets \textbf{travel} scheme (() pets (for) \textbf{short} ()) (,) (for) pets \textbf{returning} (from) abroad (to) (the) \textbf{united} kingdom (.) (this) replaced (the) \textbf{old} \textbf{system}(of) 6 months compulsory qu \#\#aran \#\#tine (for) (all) domestic pets (.) 

(in) 2001 (,) (the) eu introduced (a) passport(for) pets (.)
\end{mdframed}
Example2 : \\
Visually-grounded ratio : 5/(5+16) = 0.2381\\
BERT answered correctly : 5/5\\
XDBERT answered correctly : 5/5 

\begin{mdframed}
security forces (were) (on) \textbf{high} alert (after) (an) election campaign (in) (which) (more) (than) 1 (,) 000 \textbf{people} (,) \textbf{including} \textbf{seven} election candidates (,) (have) (been) killed (.) 

security forces (were) (on) \textbf{high} alert (after) (a) campaign marred (by) violence (.)
\end{mdframed}
Example3 : \\
Visually-grounded ratio : 8/(8+16) = 0.3333\\
BERT answered correctly : 5/5\\
XDBERT answered correctly : 5/5 

\begin{mdframed}

(in) 1979 (,) (the) leaders signed (the) egypt (-) israel peace treaty (on) (the) \textbf{white} \textbf{house} \textbf{lawn} (.) (both) president begin (and) \textbf{sad} \textbf{\#\#at} received (the) nobel peace prize (for) (their) \textbf{work} (.) (the) \textbf{two} nations (have) enjoyed peaceful relations (to) (this) \textbf{day} (.) 

(the) israel (-) egypt peace agreement (was) signed (in) 1979 (.)
\end{mdframed}
\subsection{Worsened : XDBERT underperforms BERT}
Example1 : \\
Visually-grounded ratio : 11/(11+29) = 0.2750\\
BERT answered correctly : 5/5\\
XDBERT answered correctly : 0/5 

\begin{mdframed}

jean (-) claude tri \#\#chet (,) (the) \textbf{european} central \textbf{bank} president (,) \textbf{made} (it) \textbf{clear} (,) (on) wednesday (,) (that) (he) would oppose \textbf{un} \#\#war \#\#rant \textbf{\#\#ed} political \textbf{attempts} (to) remove antonio \textbf{fa} \#\#zio (:) (the) \textbf{bank} (of) italy governor (,) engulfed (in) controversy (over) (his) handling (of) \textbf{bank} takeover bids (.)

antonio \textbf{fa} \#\#zio (is) subordinate (to) jean (-) claude tri \#\#chet (.)
\end{mdframed}
Example2 : \\
Visually-grounded ratio : 11/(11+29) = 0.4167\\
BERT answered correctly : 5/5\\
XDBERT answered correctly : 0/5 

\begin{mdframed}


(the) coastline (of) \textbf{santa} monica \textbf{bay} (is) 50 miles \textbf{long} (.)
\end{mdframed}
Example3 : \\
Visually-grounded ratio : 32/(32+55) = 0.3678\\
BERT answered correctly : 5/5\\
XDBERT answered correctly : 0/5 
\begin{mdframed}

cairo (is) (now) \textbf{home} (to) (some) 15 million \textbf{people} (-) (a) \textbf{bu} \#\#rgeon \textbf{\#\#ing} population (that) produces approximately 10 (,) 000 tonnes (of) rubbish per \textbf{day} (,) \textbf{putting} (an) enormous strain (on) \textbf{public} services (.) (in) (the) \textbf{past} 10 years (,) (the) government (has) tried \textbf{hard} (to) encourage private investment (in) (the) refuse sector (,) (but) (some) estimate \textbf{4} (,) 000 tonnes (of) waste (is) \textbf{left} \textbf{behind} every \textbf{day} (,) fest \#\#ering (in) (the) heat (as) (it) \textbf{waits} (for) \textbf{someone} (to) \textbf{clear} (it) (up) (.) (it) (is) often (the) \textbf{people} (in) (the) poor \#\#est neighbourhoods (that) (are) worst affected (.) (but) (in) (some) areas (they) (are) \textbf{fighting} \textbf{back} (.) (in) shu \#\#bra (,) \textbf{one} (of) (the) northern districts (of) (the) \textbf{city} (,) (the) residents (have) \textbf{taken} (to) (the) \textbf{streets} armed (with) \textbf{dust} \#\#pan \textbf{\#\#s} (and) \textbf{brushes} (to) \textbf{clean} (up) \textbf{public} areas (which) (have) (been) \textbf{used} (as) \textbf{public} \textbf{dump} \textbf{\#\#s} (.) 

15 million tonnes (of) rubbish (are) produced daily (in) cairo (.)
\end{mdframed}
\begin{table*}
\centering

\begin{tabular}{cccccc}
\hline
\textbf{} & \textbf{Diff. to BERT weight} & \textbf{SST-2} & \textbf{QNLI}&
\textbf{QQP} & \textbf{MNLI} \\
\hline
VL-BERT & 6.4e-3 & 90.1 & 89.5 & 88.6 & 82.9 \\
VisualBERT & 6.5e-3 & 90.3 & 88.9 & 88.4 & 82.4 \\
Oscar & 41.6e-3 & 87.3 & 50.5 & 86.6 & 77.3 \\
LXMERT & 42.0e-3 & 82.4 & 50.5 & 79.8 & 31.8\\
\hline
BERT/ViLBERT & -- & 90.3 &  89.6 & 88.4 & 82.4\\
\hline
\end{tabular}
\caption{Results of using Visual-Text Transformers on Natural Language Understanding reported by \citet{tan-bansal-2020-vokenization}. ViLBERT is identical to BERT because its weights are frozen during multimodal finetuning.}
\label{tab:VisualModelsOnGlue}
\end{table*}
\begin{table*}
\centering
\small
\begin{tabular}{ccccccccccc}
\hline
\textbf{} &\textbf{RTE} & \textbf{MPRC} & \textbf{STSB} & \textbf{CoLA}  & \textbf{SST2} & \textbf{QNLI} & \textbf{QQP} & \textbf{MNLI} &   \textbf{SWAG} & \textbf{READ↓}  \\
\hline

XBERT-b & 69.31  & 88.46 & 89.59 &  59.05 & 92.89 & 91.47 &  89.37 & 84.62 & 81.34 & --\\
XELECTRA-b & 79.78 & 91.06 & 91.46 & 66.8 & 95.06 & 93.04 & 90.62 & 88.97 & 88.91 & --\\
XELECTRA-l & 88.45 & 92.33 & 92.04 & 70.51 & 97.36 & 94.97 & 91.4 & 91.03 & 92.83 & 0.565\\
\hline
\end{tabular}
\caption{
NLU task results using the full model. 
}
\label{tab:XBERT}
\end{table*}

\begin{table*}
\centering
\begin{tabular}{c|c}
\textbf{MLM+MATCH+VC} & \textbf{MLM+MATCH} \\
\hline

\textbf{Cross-Encoder, Layer1, Head0} & \textbf{Cross-Encoder, Layer1, Head0} \\
\includegraphics[scale=0.92]{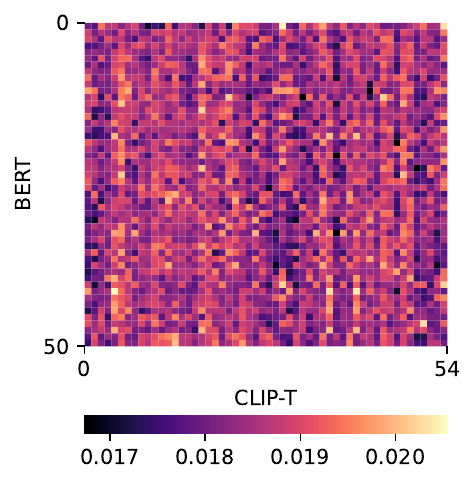} &
\includegraphics[scale=0.92]{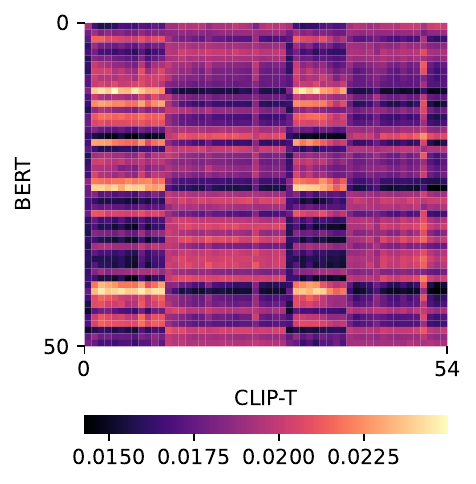} \\

\textbf{Cross-Encoder, Layer2, Head0} & \textbf{Cross-Encoder, Layer2, Head0} \\
\includegraphics[scale=0.92]{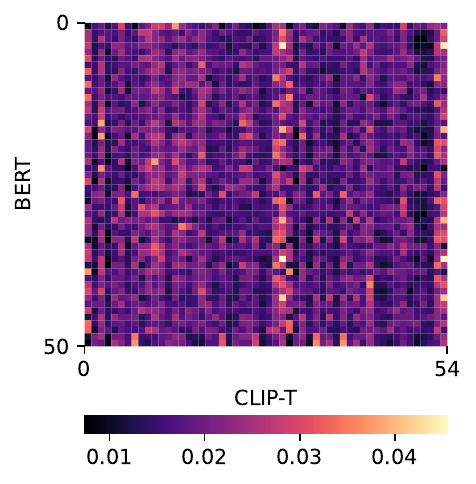} &
\includegraphics[scale=0.92]{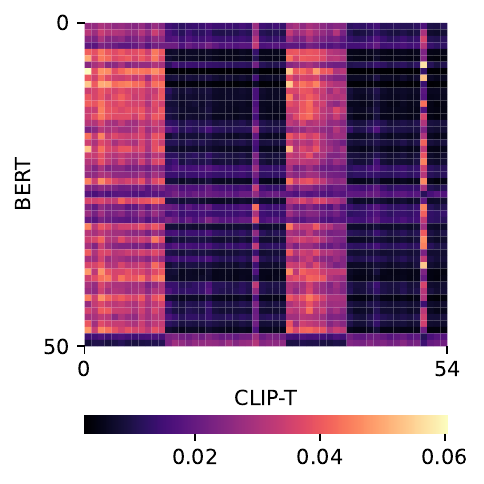} \\
\end{tabular}
\caption{Attention map of the cross-attention layers.different experiments. Left: Trained with visual classification loss, Right : Trained without visual classification loss. When trained with VC loss, the different tokens of BERT attends to the different tokens of CLIP-T more diversely. \\ BERT sequence : ['[CLS]', 'scientists', 'had', 'observed', 'that', 'mice', 'with', 'a', 'defective', 'k', '\#\#lot', '\#\#ho', 'gene', 'aged','prematurely', 'and', 'wondered', 'if', 'an', 'enhanced', 'gene', 'would', 'have', 'an', 'opposite', 'effect', '.', '[SEP]', 'scientists', 'have', 'discovered', 'a', 'gene', 'that', 'produces', 'a', 'hormone', 'that', 'raises', 'the', 'life', 'expect', '\#\#ancy', 'in', 'mice', 'by', '30', 'percent', '.', '[SEP]'] \\
CLIP-T sequence : ['<|startoftext|>', 'scientists', 'had', 'observed', 'that', 'mice', 'with', 'a', 'defe', 'ctive', 'klo', 'tho', 'gene', 'aged', 'pre', 'matu', 'rely', 'and', 'wondered', 'if', 'an', 'enhanced', 'gene', 'would', 'have', 'an', 'opposite', 'effect', '.', '<|endoftext|>','<|startoftext|>', 'scientists', 'have', 'discovered', 'a', 'gene', 'that', 'produces', 'a', 'hormone', 'that', 'raises', 'the', 'life', 'expect', 'ancy', 'in', 'mice', 'by', '3', '0', 'percent', '.', '<|endoftext|>']}
\label{tab:AttentionMap}
\end{table*}
\end{document}